# A Simple Packing Algorithm for Optimized Mapping of Artificial Neural Networks onto Non-Volatile Memory Cross-Bar Arrays

Wilfried Haensch

Materials Science Division, Argonne National Laboratory, Lemont, IL 60439

Corresponding Author: whaensch@anl.gov

**Abstract:**

Neuromorphic computing with crossbar arrays has emerged as a promising alternative to improve computing efficiency for machine learning. Previous work has focused on implementing crossbar arrays to perform basic mathematical operations. However, in this paper, we explore the impact of mapping the layers of an artificial neural network onto physical cross-bar arrays arranged in tiles across a chip. We have developed a simplified mapping algorithm to determine the number of physical tiles, with fixed optimal array dimensions, and to estimate the minimum area occupied by these tiles for a given design objective. This simplified algorithm is compared with conventional binary linear optimization, which solves the equivalent bin-packing problem. We have found that the optimum solution is not necessarily related to the minimum number of tiles; rather, it is shown to be an interaction between tile array capacity and the scaling properties of its peripheral circuits. Additionally, we have discovered that square arrays are not always the best choice for optimal mapping, and that performance optimization comes at the cost of total tile area.

# 1    Introduction

The dream of emulating the operations of the brain is the driving force behind neuromorphic computing [1] [2] [3]. Coming even close to the capabilities of the brain, however, has been elusive. The use of artificial neural networks (ANN) for machine learning is a rapidly advancing step in this direction [4] [5]. ANNs allow domain-specific learning without knowledge of the intricate details of a specific domain. Instead, they connect numerical representations of domain-specific inputs with domain-specific outputs [6] [7] [8]. A neural network, in general, has several layers represented by a weight matrix. The input signal travels through the network from layer to layer using a nonlinear activation between layers to emerge at the output as a classification of the input object. To create a model for a specific domain, a predetermined input is mapped onto itself during the training process. This is done by adjusting the weight parameters in each layer to find the minimum discrepancy between input and output. Using a finite, although possibly very large, number of such input data points one hopes for a generalization that works for the complete domain. Unfortunately, it cannot be guaranteed that this is the case [9]. Currently, the trend is to use more data and increasingly complex networks to achieve more reliable results [10] [5]. This trend stresses available computing resources, especially for building the model, although, taking advantage of certain features in the data can significantly reduce complexity and required computing resources. There are several options available to mitigate this problem from either the hardware or the software side. For instance, the use of graphic processing units (GPUs) [10] [11] or custom-made application-specific integrated circuits (ASICs) tailored for efficient execution of the most dominant operations [12] [13]. To a large extent, these operations are vector-matrix manipulations that can be computed in parallel on these hardware solutions. Furthermore, computing efficiency can be enhanced by working with reduced precision representation. The resiliency of the algorithms allows trading off computing efficiency versus numerical precision without loss of classification accuracy [14]. Based on these two observations the use of cross-bar arrays with memristor, or non-volatile memory (NVM), like elements are considered an alternative to conventional digital approaches. This could ultimately lead to significantly improved computing efficiency [15] [16] [17] [18]. In a cross-bar array configuration, shown in Figure 1, an artificial neural network $L$ that consists of $i=N_L$ layers is mapped onto a system of $N_{Tile}$ physical arrays $T_k$, or tiles, in which the arrays, of dimensions $n_k \, x \, m_k$ are embedded, with $k=1..N_{Tile}$. For practical details related to a hardware implementation, it is reasonable to assume that all arrays have the same dimension $n \, x \, m$, but are not necessarily square. The question at hand is: determining the optimal dimension of an array $n_{opt} \, x \, m_{opt}$ to minimize the chip area for a certain class of neural networks. The optimal tile area will be determined by its array capacity $n_{opt} \, x \, m_{opt}$, the array unit cell area, and the peripheral circuits needed for tile operation, as indicated in Figure 1b. The size of the array unit cell is determined by the details of the cross-bar cell and can vary a great deal [19] [20] [21] [22] [23]. The $n_{opt} \, x \, m_{opt}$ array is built by repeating the unit cell $n_{opt}$-times in vertical and $m_{opt}$-times in lateral direction, shown in Figure 1c, similar to the construct of a memory array. Therefore, the cell area, tile array capacity, and peripheral circuits will determine the tile area. To solve the mapping problem above, the details of the unit cell are not important, although they are for the resulting area. This paper does not address mapping one particular network onto an array of tiles. It addresses, rather, how networks can be mapped onto a fixed array tile configuration meeting various design objectives: such as minimum area or maximum performance, or anything in between. It is of general relevance because a chip that is optimized for only one network will not be commercially viable.

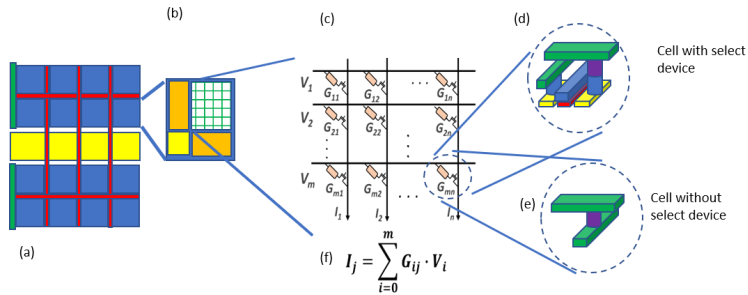

**Figure 1:** Implementation of cross-bar arrays for vector x matrix operations. (a) Chip architecture: tiles(blue), inter-tile communication(red), auxiliary digital circuits (yellow), input/output circuits (green); (b) tile architecture: control circuits (yellow), peripheral circuits(brown), array (hatched); (c) cross-bar array; (d) unit cell with select device; (e) unit cell without select device; (f) operation of vector-matrix multiplication using Ohm's and Kirchoff's law;

The implementation of analog cross-bar arrays for neuromorphic computing faces many challenges. By far the most important ones are related to the NVM material properties in which the weights are encoded [24]. Encoding can be on a continuous analog scale or in a digitized form as used in memory applications. For both usages material characteristics can impact the unit cell size or peripheral circuits [25] [26] [23] [27]. For instance, low bit resolution allows a simplified periphery but requires bit slicing to accommodate the required weight precision [24] [28]. This multiplies the number of physical tiles per network layer and will impact the chip area accordingly. Operational questions like the interconnectivity between tiles, high precision activations, which require additional computation resources, and an overall control unit will also impact the chip area due to additional digital circuitry shared among tiles. Weight transfer from digital to analog or analog to digital might require additional peripheral circuitry which might impact the tile area as well [29] [30]. In this work, we focus on mapping ANNs to physical array tiles with a given optimized capacity. This is similar to a two-dimensional bin-packing problem in which a minimum number of bins (array tiles) is found to accommodate a fixed population of input items (ANN layers). In the optimization process, we find that the number of array tiles is competing with tile capacity to minimize the utilized area. This leads to the interesting result that a minimum number of array tiles does not necessarily result in the minimum chip area because the array efficiency (defined in Equation 1 below) will scale with the array tile capacity (bin size).

This paper is organized as follows. In Section 2 we discuss mapping the layers, or weight matrices, of ANNs to a set of physical tiles with a fixed array capacity. We will show how the mapping process relates to a bin-packing problem for which binary linear optimization can be used to find an optimal solution. Further, we introduce objectives for high-density and high-performance mapping. We continue in Section 3 introducing a simplified packing algorithm and comparing it with the rigorous linear programming approach. This simplified algorithm is then used to find an optimized tile configuration, providing the main results of this work. In Section 4 a general discussion of the optimization findings and possible extensions of the process are discussed. Section 5 concludes the paper.

## 2  Mapping ANN Layers to Cross-Point Array Tiles

Mapping an artificial neural network $L$ with $N_L$ layers onto $N_{tile}$ physical tiles $T$ with a given array dimension, or capacity, is a two-dimensional bin-packing problem. The objective of this optimization is to obtain the solution that requires the smallest possible chip area. To determine this chip area, it is important to include not only the number of tiles that are required but also additional auxiliary circuitry for overall control of data flow, additional execution of high-precision arithmetic, and on-chip memory. All this would add an area $A_{aux}$ to the total tile area shown in Figure 1a. The tile area itself, Figure 1b, will contain the core array $T_{array}$ of dimension $n \times m$ built from the unit cell $D_{unit\ in} \times D_{unit\ out}$, with the vertical (in) $D_{unit\ in}$ and lateral dimension (out) $D_{unit\ out}$ dimensions aligned in $n=n_{row}$ rows (horizontal) and $m=n_{col}$ columns (vertical), a programmable control unit with area $T_{cnt}$, that stores the addresses for the inter-tile communication and synchronization $T_{cnt}=D_{cnt}^2$, and peripheral support circuitry $T_{per}$, such as analog-to-digital-converter (ADC), digital-to-analog-converter (DAC) and arithmetic engines to calculate simple activations. This will determine the tile efficiency $T_{eff}$ defined as

$$T_{eff} = \frac{T_{array}}{T_{array}+T_{cnt}+T_{per}} \tag{1}$$

$T_{eff}$ is a measure of how much of the tile area is used for weight storage. The larger this number, the better the use of tile area for weight storage. If we assume a tile layout according to Figure 1b the tile efficiency can be expressed as

$$T_{eff} = \frac{D_{unit\ in}\ D_{unit\ out}\ n_{row} n_{col}}{D_{unit\ in}\ D_{unit\ out}\ n_{row} n_{col}+(D_{unit\ in}\ n_{row}+D_{unit\ out}\ n_{col})D_{cnt}+D_{cnt}^2} \tag{2}$$

in terms of the tile array capacity $n_{row} \times n_{col}$. The tile array capacity will be an important variable in the optimization process. For bit slicing the physical mapping per layer must be multiplied by the detailed bit slicing configuration. This paper will not consider the simultaneous optimization of optimal bit slicing in conjunction with minimizing the chip area.

In the mapping process, the network layers $L_i$, $i=1..N_L$ are mapped onto the physical array configuration. The simplest solution would be to map every network layer $L_i$ to exactly one physical array $T_i$. This solution is impractical, however. Neural network layers come in a large distribution of sizes from relatively small to large which would produce suboptimal mapping. For best mapping efficiency it must be allowed that some arrays will host multiple layers while for others a network layer will be distributed over several arrays. Hardware demonstrations often circumvent this fact by using smaller networks for which the brute force 1:1 mapping – one layer per tile array – is feasible. How network layers are mapped to physical arrays will determine if the execution can be pipelined or not. Pipelining means the simultaneous use of all network layers for maximum performance throughput. This means that all arrays, that contain either segments or multiple layers of the neural network, will also be activated at the same time. Especially, if an array includes several complete or partial layers it can come to undesired interference between shared input and output of the layers which will compromise the output signal. We show in Figure 2 mapping scenarios of network layers onto physical array tiles. The left, Figure 2a, and middle, Figure 2b, arrays have shared input or output lines across the network layers which prohibits pipelining because either inputs or outputs are mixed across network layers. Pipelining requires non-overlapping input and output channels to be a constraint in the mapping process as shown in Figure 2c. This comes as we will show, however, at the cost of chip area. This further aggravated for convolutional neural networks (CNNs) as we will discuss later due to the large weight reuse factor in these networks.

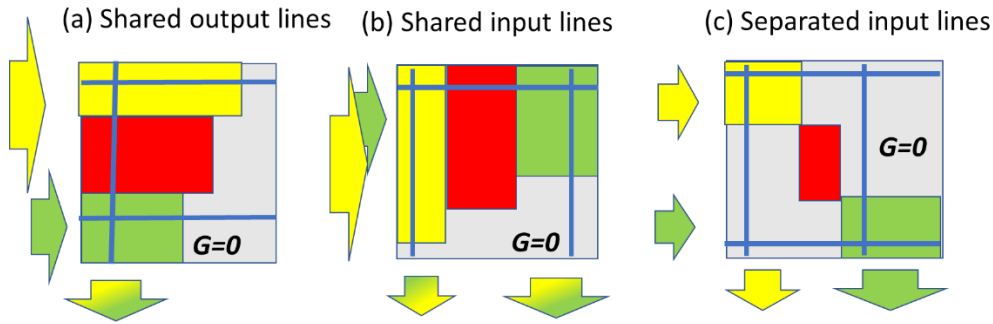

(a) Shared output lines   (b) Shared input lines   (c) Separated input lines

**Figure 2:** Mapping of network layers (yellow, red, green) onto physical arrays(grey). All vertical and horizontal lines of the physical array are operated simultaneously through the peripheral circuits. Input is received on the horizontal lines and the output is collected on the vertical lines. If network layers share an output line (a) the output signal will collect contributions from different network layers. If network layers share an input line (b) the input will be distributed across multiple network layers which will result in a mixed output as well. Only if no output and input lines are shared (c) can a clear network layer-specific signal be obtained at the output. Only in this latter case can all logic layers be operated simultaneously. Cross-points in the grey area are not assigned to any ANN layer and are programmed to have very low conductance *(G=0)*

For a sequential execution, in general, only one layer per tile is activated at a time and the signal has to travel through all network layers before the next signal can be launched. The execution time per layer is $t_k$ The latency will be,

$$t_{latency} = \sum_k t_k + t_{dig} + t_{com} = t_{tile} \sum_k N_{reuse}^k + t_{dig} + t_{com} \tag{3}$$

The index *k* runs over all layers with $t_{tile}$ the execution time per tile, $N_{reuse}^k$ the weight reuse factor of layer *k* ($t_k = t_{tile} N_{reuse}^k$) and $t_{dig}$ and $t_{com}$, are times associated with additional digital processing and communication, respectively. The tile execution time $t_{tile}$ contains the integration time $t_{int}$ of the charge on the output lines, the conversion of this charge into a digital signal by the ADC, and the calculation of a simple activation function. Properly designed the latter two are hidden and the tile execution time is close to the integration time. $t_{tile} \approx t_{int}$. Additional digital operations $t_{dig}$, for instance for more complex activations or non-analog data manipulations and communication between tiles $t_{com}$, are hidden as well so that the total execution time can be estimated by the single tile integration time, the number of layers and their weight reuse. The tile execution time will be the same for all tiles to allow synchronous execution to move signals from one tile to the next in a systematically clocked way.

For pipelined execution, the latency is given by the slowest of the components (layer, communication, additional digital processing) of the network,

$$t_{latency} = \max(t_{tile} N_{reuse}^{max}, t_{com}, t_{dig}) \tag{4}$$

Equation 3 implies that the latency of a non-pipelined implementation for networks with only fully connected layers ($N_{reuse}^k = 1, \sum N_{reuse}^k = N_L$) will increase with the number of network layers. For CNNs, the latency will increase even faster because of the high weight reuse per layer. CNNs were originally introduced to reduce the number of weight parameters for image recognition [7], but have wider use in spatially correlated data. A large weight reuse factor in CNN layers implies that pipelined implementations are limited by the tile processing time multiplied by the maximal weight reuse, Equation 4. Therefore, the weight reuse becomes a major performance detractor for pipelining. Typically, in CCNs, most of the execution time is spent in the first layers due to the large weight reuse. Therefore, significant performance gain between pipelined and non-pipelined implementations is moderate in cross-bar architectures. To address this problem further parallel processing can be pursued by weight cloning. Figure 3 shows the basic idea for this cloning [31], or RAPA (Replicated Arrays with Permuted Assignment) process. The convolution process can be converted into a simple matrix-matrix multiplication *IM x WM*. The matrix *WM* contains the weights of the convolution filters and *IM* is constructed by column vectors, derived from the input data, so that the convolution is correctly processed. The height of *IM* is $k^2 d_{in}$ *(+1)* (the *(+1)* addition is needed if an activation bias is used) and its width $((n_{in}-k+2p)/s+1)^2$, with *k* the filter kernel dimension, *p* the padding, *s* the stride, $n_{in}$ the input dimension, and $d_{in}$ the number of input channels. The size of *WM* is given $d_{out}$ *x* ($k^2 d_{in}$*(+1)*) with $d_{out}$ the number of output channels.

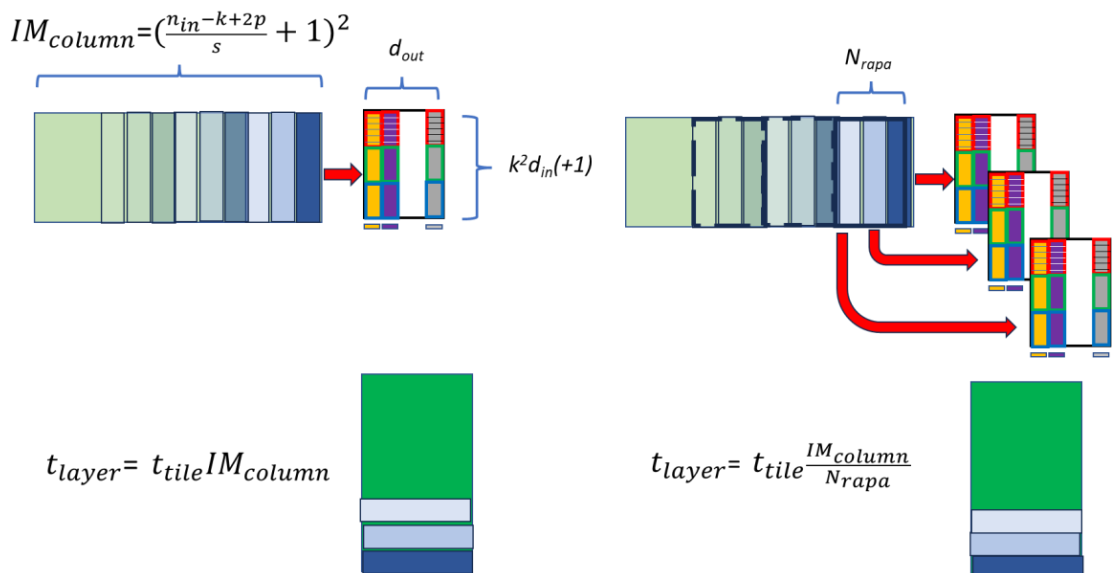

**Figure 3:** RAPA (*Replicated Arrays with Permuted Assignment*) expedites CNN. Left-hand side: In a conventional convolution the input matrix (upper rectangle) is column-wise multiplied with the weight matrix producing an output matrix of the same size (lower rectangle), however, transposed. Each column x weight matrix multiplication is done in $t_{tile}$ given the total layer

processing time $t_{layer}$ as a multiple of *IM* columns $IM_{column}$, $IM_{colums}$ is the weight reuse factor of the layer. Right-hand side: for RAPA the weight matrix is replicated $N_{rapa}$ times. This enables $N_{RAPA}$ columns of the input matrix to be processed in parallel, therefore, reducing the layer processing time by $N_{RAPA}$. Network parameters stride s, filter kernel dimension k, and padding p control the convolution. The incoming data has the dimension $n_{in} \times n_{in}$.

The network design parameters $n_{in}$, *s, k,* and *p* determine the number of columns of *IM*. This number will approximately quadratically increase with the size of the input dimension $n_{in}$ and therefore the performance penalty will be significantly larger if larger input dimensions are considered. In Table 1 we show the weight reuse for the first convolution layer for selected convolutional neural networks with their standard parameters.

**Table 1:** Weight reuse for selected CNN

| Network | ResNet50 | ResNet9 | AlexNet [11] | LeNet [7] |
|---|---|---|---|---|
| Input (#Images) | ImageNet (1.2M) | Cifar10 (60k) | ImageNet | MNIST (60k) |
| Input size (pixels) | 3 x 224 x 224 | 3 x 32 x 32 | 3 x 224 x 224 | 1 x 28 x 28 |
| $N_{reuse}$ 1st layer | 12544 | 729 | 3025 | 784 |

The ResNet architecture is described in [32], where the number refers to the layers in the network. The dataset MNIST [33] provides 8-bit grayscale $28 \times 28$-pixel images of handwritten digits, CIFAR10 [34] consists of $32 \times 32$-pixel, low-contrast, red-green-blue (RGB) color maps with 10 classification categories and ImageNet [35] is a very large dataset of 224x224-pixel, low contrast, red-green-blue (RGB) color maps with 1000 classification categories. The impact of weight reuse for larger input image sizes and more complex networks is seen. To mitigate the effect of large weight reuse the replication of the weight matrix *WM* can be used to reduce the layer processing time, as shown in Figure 3 (right side: RAPA) [31]. All in all, the replication must be chosen layer by layer so that the computational load is similar across the network to achieve load balance. Otherwise, the slowest layer will be the performance bottleneck. In addition, to ensure pipelined processing of the replicated weight matrix they must be mapped onto a non-overlapping configuration to the physical arrays (Figure 2c) as discussed above.

## 2.1 Connecting the Mapping Process to the Bin-Packing Problem

We will now formulate the network-to-array tile mapping in an appropriate manner suitable for a bin-packing problem so that we can use bin-packing algorithms to determine the optimal placement of network layers onto a population of array tiles. A bin-packing problem needs the input items and the bins in which the input items are packed. The input items are the artificial neural network layers and the bins are the physical arrays onto which the network layers are mapped. The physical array is associated with the tile array of dimension $T(n_{row}, n_{col})$ which does not necessarily need to be square ($n_{row}=n_{col}$). The required number of tiles for a minimum total tile area will be the result of the optimization.

The network $L$ will have $N_L$ layers of weight matrices with dimension $m_{inp,i}$ and $m_{out,i}$, or $L_i(m_{inp,i}, m_{out,i}), i=1..N_L$. If the network weight matrix $L_i$ is larger than the physical array, $m_{inp,i}, m_{out,i} > n_{row}, n_{col}$, the layer has to be distributed across several array tiles, or fragmented into $k1$ fully mapped arrays $T_i^j$, $j=1..k1$, plus the rest $T_i^{k1+l}$, $l=1..k2$ which are arrays that still have space available. For the network layer $L_i$ we therefore have

$$L_i^j(n_{row}, n_{col}); j = 1..k1_i \tag{5a}$$

$$L_i^{k1+l}(q_{inp,i}, q_{out,i}); q_{inp,i}, q_{out,i} < n_{row}, n_{col}; l = 1..k2_i \tag{5b}$$

The fragmentation will produce a new set of logical blocks $LF$, associated with the neural network layer $L_i$, $FL_i^j$, $j=1..k$, $k=k1_i+k2_i$, with input dimensions $p_{inp,j} \leq n_{row}$ and $p_{out,j} \leq n_{col}$. It will turn out to be useful to sort these new blocks in descending order of row dimension as input for a simplified bin-packing algorithm that we introduce below. The fragmentation process will of course produce a different set of logical blocks for each physical array dimension. As an example, we show in Figure 4 the fragmentation of ResNet18 on ImageNet for different square physical arrays.

The fragmentation produces four different kinds of blocks:

i) Fully mapped arrays ($p_{in}=n_{row}$, $p_{out}=n_{col}$)
ii) Row dimension is fully mapped ($p_{in}=n_{row}$, $p_{out}<n_{col}$)
iii) Column dimension is fully mapped ($p_{in}<n_{row}$, $p_{out}=n_{col}$)
iv) Sparse - neither row nor column dimension is fully mapped ($p_{in}<n_{row}$, $p_{ou}<n_{col}$)

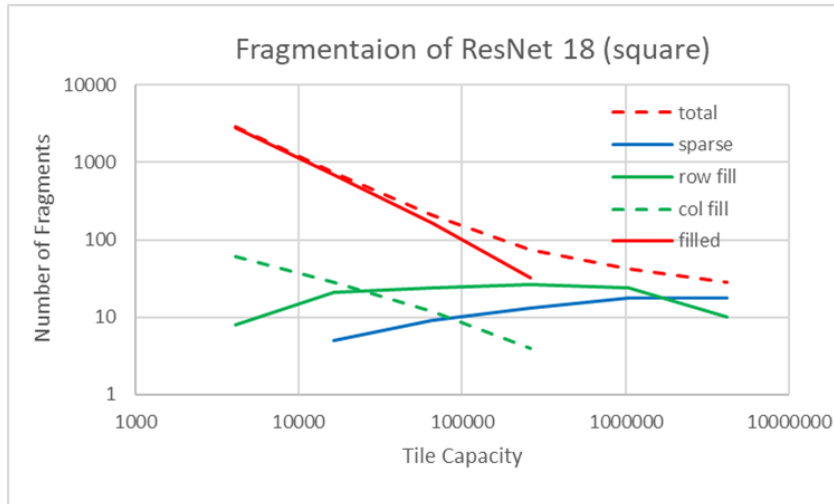

**Figure 4:** Fragmentation of ResNet18 on ImageNet onto square arrays. Total number of fragmented blocks (red dash); blocks that are fully mapped (red solid); blocks that are fully mapped in row dimension (green solid); blocks that are fully mapped in column dimension

(green dash); residual blocks (blue solid); The difference in row-fill and column-fill is due to the aspect ratio of the logical layers which are generally not square.

Scenario i) has a filled physical array, there is no more available space. In ii) and iii) either the row dimension or the column dimension is fully mapped. In either case, space is available in these arrays for dense packing. However, no space is available for pipeline packing since it would lead to shared input or output. Case iv) opens the possibility for shared space with blocks from other layers. From an optimization point of view for pipeline mapping only blocks from case iv) need to be considered while for dense packing cases ii), iii), and iv) need to be considered simultaneously to find the optimal solution to minimize the total tile area.

We now have all the necessary components to implement a bin-packing algorithm with the fragmented network (input) and the corresponding physical array size (bins). It's important to note that each choice of array size will lead to a distinct fragmentation of the neural network across the array tiles, resulting in a different mapping of the neural network. Before we address the optimized mapping problem, we will provide a brief discussion of bin-packing and its relation to the mapping problem in the next section

## 2.2  Linear Optimization to Solve Bin-Packing

The bin-packing problem can be solved by linear optimization. This will not provide the exact solution to the problem but will find a mathematical optimum. Algorithms of this kind are used for a variety of purposes for instance delivery schedules, warehouse management, and more. For our purposes, we will use the LPSsolve (LPS) package [36], which allows optimization across floating point, mixed-integer, and binary (assuming only the values 0 and 1) variables. LPS can be operated as a standalone, called from within program environments, or used in an integrated fashion with several programming tools. In this work, LPS is used as part of MAPLE [37] and as a standalone application. Typically, these algorithms start with an allowed maximum number of bins $N_{max}$ (typically chosen to be the number of items to be packed) and a successful optimization will then produce the number of bins $N$ that is hopefully significantly lower than the given maximum.

According to our discussion above we have to distinguish two packing scenarios: first, a scenario where inputs or outputs are shared (Figure 2a and Figure 2b) and second, where neither outputs nor inputs are shared (Figure 2c). The former will give the highest density and the latter will result in the highest performance due to the ability to pipeline. Both dense packing and pipeline packing will require separate algorithms, which are described now.

### 2.2.1 Dense Packing

Table 2 defines the variables for the dense packing algorithm [38]

**Table 2:** Variable definition for dense packing

| Variable | value | Purpose | comment |
|----------|-------|---------|---------|
| y[i] | 0,1 | =1: Item i initialized the first layer in the first bin | always start in the left lower corner (1,1) |
| q[i] | 0,1 | =1: First layer in bin i | sum over q is the number of used bins |
| z[i,j] | 0,1 | =1: Item j initialized layer in bin i | Filling up bin i |
| x[i,j] | 0,1 | =1: Item j goes in the layer initialized by item i | Filling up the layer initialized by item i |
| LF[i,1] | input | Row dimension of i-th fragmented block | |
| LF[I,2] | input | Column dimension of i-th fragmented block | |
| T[1] | input | Row dimension of the physical tile array | |
| T[2] | input | Column dimension of physical tile array | |

The optimization problem is formulated as:

$$N = \sum_{i=1}^{N_{max}} q[i] \qquad (6a)$$

$$\sum_{i=1}^{j-1} x[i,j] + y[j] = 1, j = 1..N_{max} \qquad (6b)$$

$$\sum_{j=1}^{i+1} LF[j,1]x[i,j] - (T[1] - LF[i,1])y[i] \le 0, i = 1..N_{max} \qquad (6c)$$

$$\sum_{k=1}^{i-1} z[k,i] + q[i] - y[i] = 0, i = 1..N_{max} \qquad \text{Equ. 6e}$$

$$\sum_{i=k+1}^{N_{max}} LF[i,2]z[k,i] - (T[2] - LF[k,2])q[k] \le 0, k = 1..N_{max} - 1 \qquad (6d)$$

The number of used bins *N*, Equation. 6a, is minimized under the constraints of Equations 6b to 6d with $N_{max}$ set to the number of components in *LF*. Since the independent variables can only assume the values 0 or 1 the optimization is done in binary variables using a branch-and-bound algorithm. To visualize the process we will, for demonstration, use a simple bin-packing problem. A list $FL=\{FL_i\}$ of $i=1..13$ components, represented by

$FL= \{(257, 256), (257, 256), (257, 256), (129, 256), (129, 128), (129, 128), (129, 128), (65, 128), (148, 64), (65, 64), (65, 64), (65, 64), (65, 64)\}$ (7)

is mapped on physical arrays $T$ of fixed capacity $T(512,512)$. For simplicity, we label the components of the list Items 1 to Item 13 in successive order as they appear in the list. The result of linear programming under the constraints of Equation 6 is shown in Table 3 and visually depicted in Figure 5.

Table 3: Bin-packing for dense packing

|         | Bin 1   |        |         |  | Bin 2  |         |         |         |
|---------|---------|--------|---------|--|--------|---------|---------|---------|
| **Layer 1** | Item 1  | Item 4 | Item 11 |  | Item 2 | Item 10 | Item 12 | Item 13 |
| **Layer 2** | Item 5  | Item 6 | Item 9  |  | Item 3 | Item 8  |         |         |
| **Layer 3** | Item 7  |        |         |  |        |         |         |         |

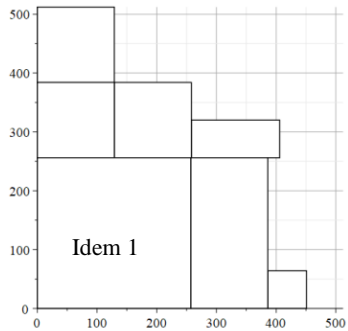
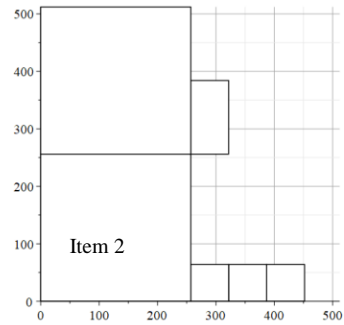

Bin 1              Bin 2

**Figure 5:** Graphical representation of dense bin-packing problem Equation 7 with a $T(512,512)$ physical array.

We find that the optimal solution requires 2 bins or two arrays of the given capacity.

### 2.2.2 Pipeline Packing

We now consider the same problem for pipeline packing. The optimization formulation is different from the dense packing because now we stack non-overlapping components to ensure input and output do not interfere. In contrast to the dense packing, there will be only one component per layer [39] which reduces the problem to an effective one-dimensional packing problem. The variables are presented in Table 4.

**Table 4**: Variable definition for pipeline packing

| Variable | value | Purpose | comment |
|---|---|---|---|
| y[j] | 0,1 | =1: bin j is used | |
| x[i,j] | 0,1 | =1: item I is packed in bin j | |
| LF[i,1] | input | Row dimension of i-th fragmented block | |
| LF[I,2] | input | Column dimension of i-th fragmented block | |
| T[1] | input | Row dimension of physical tile | |
| T[2] | input | Column dimension of physical tile | |

The modified optimization problem is formulated as,

$$N = \sum_{i=1}^{N_{max}} y[i] \tag{7a}$$

$$\sum_{i=1,j=1}^{N_{max},N_{max}} x[i,j] = 1 \tag{7b}$$

$$\sum_{i=1}^{N_{max}} LF[i,1]x[i,j] - T[1]y[j] \leq 0, j = 1..N_{max} \tag{7c}$$

$$x[i,j] \leq y[j], i = 1..N_{max}, j = 1..N_{max} \tag{7e}$$

$$\sum_{i=1}^{N_{max}} LF[i,2]x[i,j] - T[2]y[j] \leq 0, j = 1..N_{max} \tag{7d}$$

again, to find the minimum number $N$ of bins. For the same problem as above, we now obtain bin-packing in 4 bins with non-overlapping shapes shown in Table 5 and Figure 6.

**Table 5:** Bin-packing for pipeline packing

| Bin 1 | Bin 2 | Bin 3 | Bin 4 |
|---|---|---|---|
| | | | |

| Item 3 | Item 2 | Item 5 | Item1 |
| --- | --- | --- | --- |
| Item 4 | Item 9 | Item 6 | Item 8 |
|  | Item 13 | Item 7 | Item 10 |
|  |  | Item 11 | Item 12 |

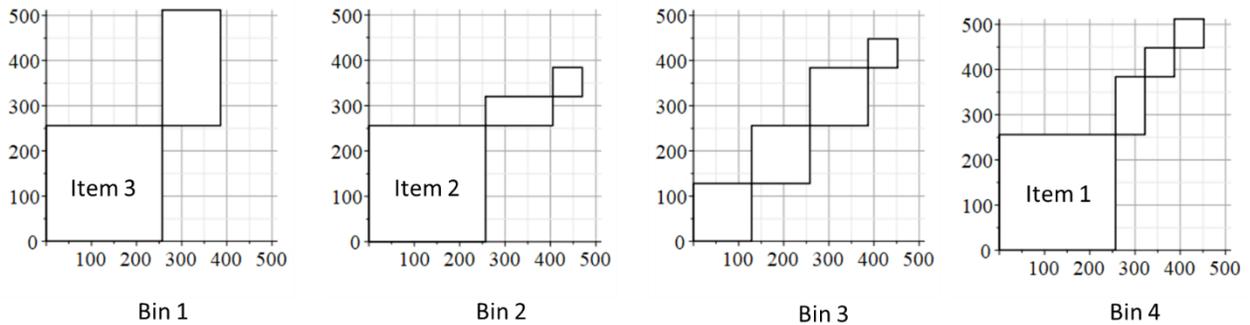

**Figure 6:** Graphical representation of pipeline bin-packing problem Equation 7 with a *T(512,512)* physical array.

Comparing Figure 5 and Figure 6 we see the dramatic effect of pipeline-enabled packing. The number of used arrays (bins) goes from 2 to 4 in this simple case. For our example, we have shown the packing solution for one physical array dimension only. To find the optimal mapping strategy the bin-packing algorithm needs to cycle through many array configurations to find the optimal one. Unfortunately, the branch-and-bound algorithm increases in complexity with the increase in problem size, which is relevant for larger networks. Although, in principle, the algorithm will always find an optimal solution, if one exists, it comes at exponentially increased execution time for larger problems. Therefore, to obtain a solution is not always feasible. Often the situation can be improved by a new choice of control parameters for the algorithm. For an optimization cycle in which a large amount of array configurations is investigated, this is an unfortunate situation. It would inevitably force an individual approach for each configuration with customized control parameters.

## 3   The Simplified Bin-Packing Algorithm

Although linear programming is the method of choice, one must ask: is there an alternative that does not suffer from the above-mentioned short shortcomings? A method that will allow an investigation of a large number of array configurations producing a result on par, in quality, with the linear programming approach. We believe we have found a close approximation to the rigorous linear programming approach. We first generate a set of $N_{array}$ array configurations $\{T_j\}$ $j=1..N_{array}$ with different row and column dimensions. The elements of this set could either be square $n_{row}=n_{col}$ or rectangular with either $n_{row}> n_{col}$ or $n_{row}<n_{col}$. Secondly, we generate a fragmentation of the artificial neural net for each of the different array configurations resulting in $\{FL^j\}$, $j=1..N_{array}$ fragmented sets. There is one fragmentation $FL^j$ for each element of the set $\{T_j\}$. The elements of $FL^j$ must be ordered in ascending order of row dimension. We now have paired an array configuration $T_j$ with a fragmentation $FL^j$. Please note that the components of $FL^j$ are the fragmented layers of the neural network for the tile dimensions of $T_j$. The third and final step is to place the ordered elements into the arrays in sequence as they appear in the ordered list: the first element goes in the lower left corner of the first array and the other elements are added until the first layer is filled. Then a second layer is added starting from the left. When the first array is filled the second array is started again at the lower left corner. This will produce a layer structure similar to that shown in Figure 5 for dense packing and a staircase structure similar to Figure 6 for pipeline packing. The process is repeated for all members of the list $FL^j$ until every member is placed in several arrays of size $T_j$. Keeping track of the number of arrays used for each member completes this task. This process is straightforward, does not suffer from possible convergency problems, and is easily implemented. The question remains, however: how does this packing approach compare with the more rigorous branch-and-bound algorithm? In Figure 7,

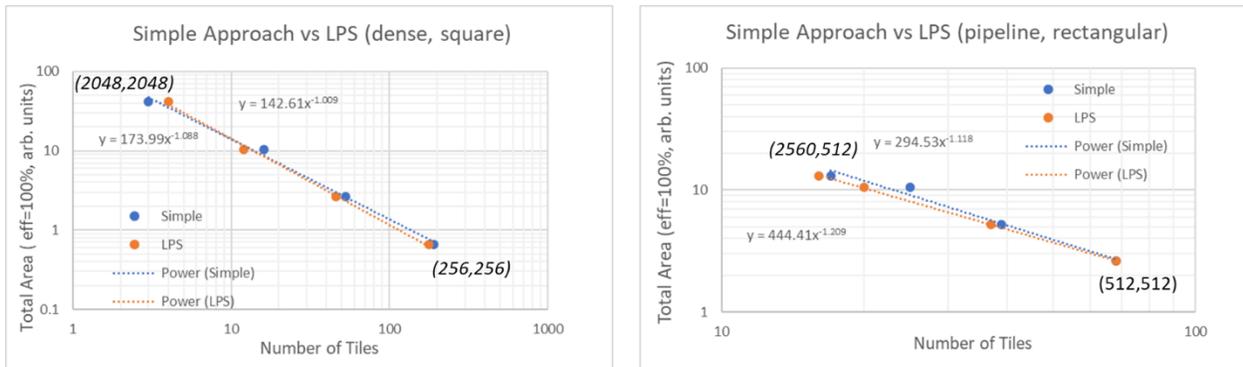

**Figure 7:** Comparison of simple packing algorithm with linear programming. The graph shows the minimum total tile area vs the number of tiles for this minimum. Tile sizes are optimization results that are not the same in both plots. The fit function for both is included in the graph. Blue circles: simple packing, brown circles: linear programming. The numbers in parentheses indicate the array dimension. Total tile area equals number of tiles x array area (100% efficiency)

For example, we compare the simple packing approach and the linear programming optimization for two scenarios: dense packing with square arrays and pipeline packing with rectangular arrays. In both cases, we use ResNet18 on ImageNet as the neural network of choice. The simplified approach shows a good correlation with the linear programming results, Figure 7. Since it captures the trend correctly, we will use the simplified approach to investigate the optimal tile capacity.

## 3.1 The Optimization Process

The optimization is a three-step process: First, create a sample of tiles with eight different aspect ratios, second, we will find the minimum total tile area for each sub-sample with a fixed aspect ratio and the corresponding number of tiles, and third, the minimum area across all sub-samples, together with its tile capacity, is the desired optimization result. To this end, we will vary the tile array row dimensions $n_{row} = 2^{5+k}$, $k=1..8$, and the tile array column dimension $n_{row}/n_{col}=i, i=1..8$ for the aspect ratio of the array. This covers the array range $T(64,64) ... T(8192,65536)$. For each aspect ratio, we record the tile with the minimum tile area which gives a list of 8 choices. The minimum tile area in that list is the optimum tile for the network under consideration. In Figure 8 we show, as an example, two optimization results. Both are done on the ResNet18/ImageNet combination on square arrays. The first, on the left-hand side, shows the optimization for a dense minimum area target, and the second, on the right-hand side, shows an area optimization for a pipeline solution. The area cost of the pipeline solution is about twice that of the dense solution. The dense optimization requires 16 tiles of dimension 1024 x 1024, while the pipeline solution will need 68 tiles at 512 x 512 array dimension. The area penalty of the pipeline solution can be cut approximately in half with 17 rectangular arrays of dimension 2560 x 512 (obtained in a separate optimization not shown). This indicates that square arrays might not be the best option for optimal area use. This improvement is gained by the better packing efficiency with rectangular arrays for this type of network.

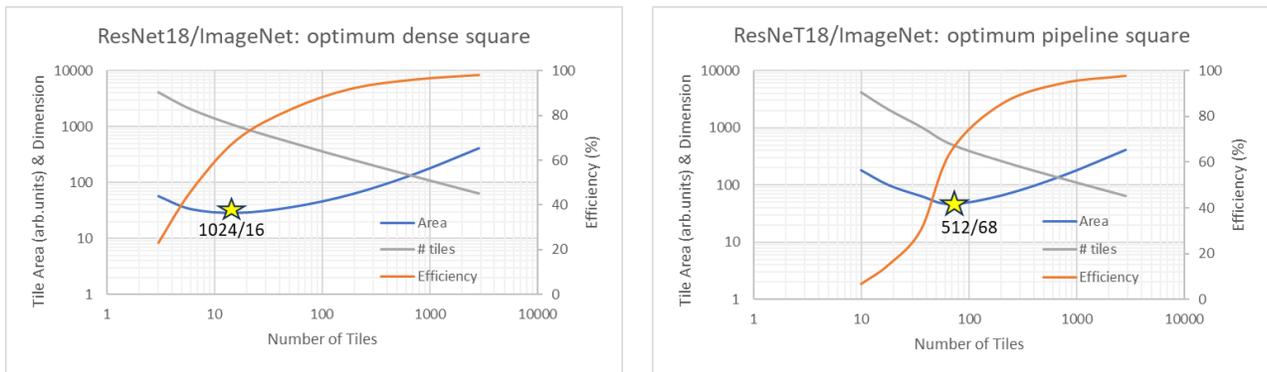

**Figure 8:** Mapping optimization of ResNet18/ImageNet on square arrays. The plot shows the minimum total tile area versus the corresponding number of tiles for various tile aspect ratios. Tile area (blue). Tile area is not chip area auxiliary components like CPU, I/O, and memory need to be added; Mapping efficiency (brown); Tile Dimension (grey). Left-hand side: dense packing for minimum tile area; Right-hand side: pipeline packing for optimum area. The yellow star indicates optimal packing: tile (array) capacity/number of tiles

We must note that a minimum number of tiles does not necessarily correspond to a minimum total tile area because the array efficiency is not constant. The array efficiency will scale with the array size of the tile. The detailed scaling behavior will depend on the base technology and design choices for the peripheral circuits. For this paper, we assume that the control circuit area, yellow square in Figure 1b, will be approximately constant because it must contain control circuits for routing the signals and a small memory that holds the information for the routing. From the literature, we can infer a tile efficiency for a certain tile size and from that the control area $D_{cnt}^2$ according to Equation 2. Further, we assume that the so-determined area for the peripherical circuits, a brown area in Figure 1b, can accommodate DAC, ADCs, and small arithmetic units to calculate simple activations. Design choices could include the increase of shared columns per ADC for instance to let the number of ADCs grow sublinear with the array dimension. Specific design choices can be easily integrated into the algorithm by accommodating how tile size scales with the array capacity. For our calculations, we use a tile efficiency of 20% at a 256 x 256 tile dimension [26]. It appears that the optimal array capacity does not strongly depend on the array efficiency but rather on the array dimension itself. The tile area, however, does. A summary of the different optimization choices for ResNet18/ImageNet w shown in Figure 9.

Figure 9 shows the clear tradeoff between area and performance concerning network mapping. It also shows the interesting result that the aspect ratio of the array $n_{row}/n_{col}$ has a significant effect on the number of tiles needed. The three groups: dense, pipeline, and RAPA have comparable area requirements at their optimum but significantly different tile counts for tiles that accommodate arrays with an aspect ratio different from one. A reduced tile count could be a benefit at the system design level. The RAPA solution has a throughput improvement of approximately a factor of 100 at the area cost of a factor of five. The improvement is even more dramatic if the performance improvement is measured against the non-pipeline dense mapping. We will discuss this further below.

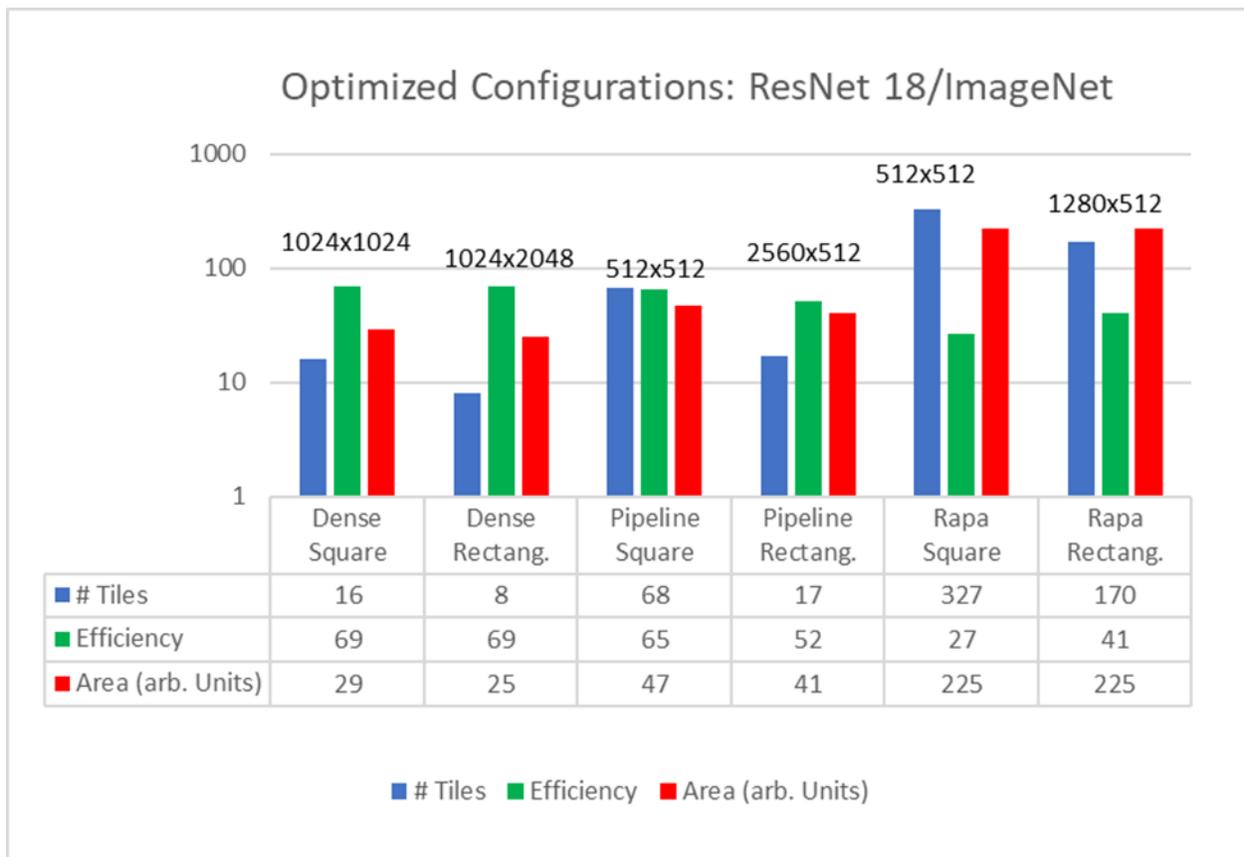

**Figure 9:** Optimum mapping configurations for ResNet/ImageNet. Number of Tiles (blue); Tile efficiency (green); Tille area arbitrary units (red); The group of bars shows from left to right: Dense square arrays, no pipeline; Dense rectangular arrays, no pipeline; Pipeline square arrays; Pipeline rectangular arrays; RAPA pipeline square arrays; RAPA pipeline rectangular arrays; $N_{rapa}$ =128 for 1$^{st}$ layer and successive reduction by 4 for balanced loading throughout the network.

The impact of the optimization on network and data complexity is shown in Table 6. For example, ResNet18 /ImageNet has 11.5M weight parameters and ResNet9/Cifar10 has 1.9M parameters. Mapping 1:1, which means each fragmented layer component has its physical tile gives the worst possible solution at a fixed array size (Mapping 1:1 - 2$^{nd}$ row). We find this is close to the mapping used in [26]. A reduction in the number of tiles is achieved by optimized packing, for both the linear programming solution (LPS - 3$^{rd}$ row) and the simplified approach (Simple approach - 4$^{th}$ row).

**Table 6:** Lage vs small Networks (dense, square)

| Array | ResNet18/ImageNet | ResNet9/Cifar10 | option |
|---|---|---|---|
| 256 x 256 | 208 (239 mm$^2$) | 40 (50mm$^2$) | Mapping 1:1 |
| | 177 (203 mm$^2$) | 34 (33 mm$^2$) | LPS |
| | 191 (219 mm$^2$) | 35 (36 mm$^2$) | Simple approach |
| 1024x1024 | 16 (66 mm$^2$) | 3 (1.2 mm$^2$) | LPS/Simple approach |

Please note that the LPS solution has a lower array count than the simplified approach, as expected. The entries in parenthesis estimate the total tile area assuming a cell design discussed in [26] showing the benefit of larger array sizes for a more efficient mapping. (LPS and the simple approach give the same result for the large array size – 5$^{th}$ row). This example is an illustration of area impact due to network complexity, data complexity, and mapping strategy.

## 4  Discussion

In the previous section, we discussed the impact of the mapping strategy with a simplified mapping algorithm. The optimum packing point, resulting in the smallest total tile area, is given by the restrictions of the network (layer sizes and shapes) and the physical array dimensions of the tiles. In other words, the bin-packing problem is the response to tile array capacity versus network layer dimensions. The impact on the total tile area or chip area will depend on detailed tile design decisions. Assuming certain scaling behavior of tile circuitry we find that a minimum number of tiles does not necessarily result in a smaller total tile area. A careful study of these tile circuits and their scalability with tile array capacity is an important ingredient for packing optimization. It is important not to confuse packing efficiency with array efficiency. The latter is the result of design choices, while the former is the result of packing objectives. Lower packing efficiency does not necessarily mean sub-optimal mapping. The optimum depends on the design objective! There is a clear tradeoff between packing density and performance. To make this point clear we employed a simple array efficiency scaling model that can be refined according to the actual design details. However, an optimum packing configuration, as far as the array capacity is concerned, is not impacted by these design choices if a monotonic scaling law for the tile circuits can be found. The simplified model can correctly reproduce the packing trend for an optimal solution without running into the uncertainties of the more elaborate binary linear optimization. Once this optimal solution is found, however, it is advisable to compare the results with the linear optimization. This would require finding the solution for only one tile configuration.

For small networks and small input data, the benefit of optimization is small if a 1:1 approach (each layer, appropriately fragmented, can be mapped to an individual tile) is possible. In that case, pipelining is possible because the network layers are perfectly decoupled. For larger networks the situation changes. In Figure 10 we compare ResNet50 on ImageNet and an individual layer of BERT [40] with 12 heads, a sequence length S of 64, and an embedding dimension of 768 using a 1:1 approach and the optimized version for pipeline implementations.

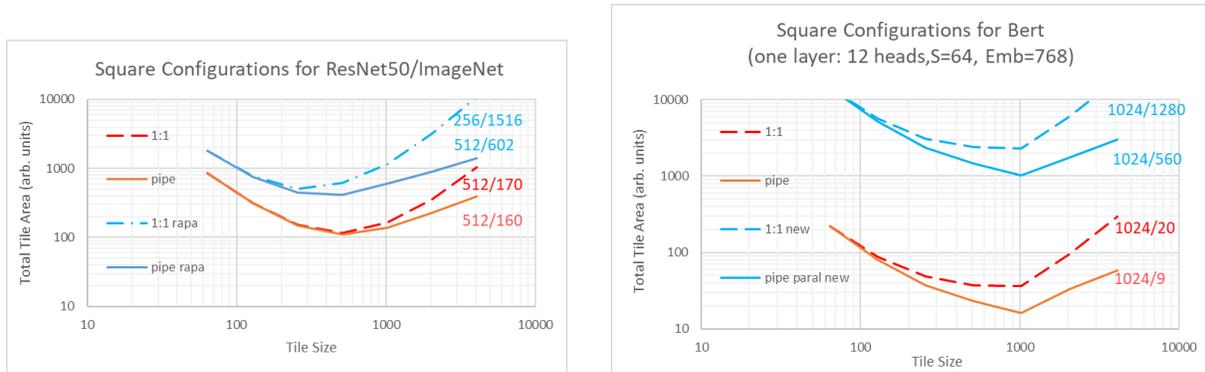

**Figure 10:** Packing optimization for square arrays. Left-hand side: ResNet50/ImageNet- mappings done with correctly fragmented network; optimization – red solid line; 1:1 – red dashed line; RAPA 128/4 – blue solid; RAPA 128/4 – blue dash-dot-dash; Right-hand side: BERT one layer- mappings done with correctly fragmented network; optimization – red solid line; 1:1 – red dashed line; Maximum parallelism – blue solid; Maximum parallelism – blue dash-dot-dash; Numbers in figure correspond to tile (array) size/number of tiles at optimum packing point. The total tile area is in arbitrary units for relative comparison.

First, we compare a 1:1 mapping of the properly fragmented network with that of mapping to non-overlapping layer fragments. Second, we utilize weight layer replication for improved throughput. Here we use for the convolutional network ResNet50 the RAPA process with an initial replication of 128 that is successively reduced by a factor of 4 (we use the notation 128/4) to achieve load balance across the network and for BERT we replicate the fully connected layers by the sequence length S. Again, we compare the 1:1 mapping with that of mapping to nonoverlapping layer-fragments. The major structural difference is the number of layer fragments. In both cases, the replication process increases the number of layer fragments significantly. The 1:1 implementation loses out at larger tile sizes (array capacity). For ResNet50 simple pipelining 1:1 and optimizing give similar results for the optimum. The situation changes for the throughput-optimized RAPA mapping; for which optimizing shows a clear advantage in the number of tiles at a similar total tile area at the optimal mapping. This is similar to the earlier observation for ResNet18 (Figure 9). The situation is quite different for BERT. Optimizing gives an improved total tile area for both simple pipelining and throughput-optimized mapping. The reason for this is the different structures of the network's weight matrixes. For the convolutional network replication is higher at the relatively small layers at the beginning and is reduced at the later layers while for BERT the replication carries the same factor throughout. The example we have shown uses square arrays. Further improvement can be obtained by allowing a variation in the aspect ratio of the tile array as indicated in Figure 9. These examples clearly show that neuromorphic computing with cross-bar arrays comes at the price of real estate, especially for modern transformer-like networks that are the backbone of large language models like ChatGPT. While for some of the smaller networks, that use less data, it might be possible to find a one-chip solution. For the larger ones most certainly multiple chips need to be used. This creates another impediment to performance and energy because signals now have to travel across chip boundaries. That is a problem that conventional digital implementation has to deal with as well.

Since the weight encoding process (write) is usually very energy and time-intensive neuromorphic systems tend not to be virtualizable without significant energy and performance penalty. This means, that when high performance and low energy is a design objective the network needs to be placed on several chips to hold all the weight simulations. Even for the smaller networks the lack of virtualization will impact the chip area, compared with a digital solution; where weights can be swapped out frequently with the reuse of compute resources in a limited chip area at the cost of memory access. This is mitigated, of course, by an on-chip cache or scratchpad. For applications with moderate performance requirements dense, non-pipeline, packing might be the best application space for neuromorphic computation with cross-bar arrays could take advantage of a possible reduced power budget and the non-volatile storage of the weights.

## 5    Conclusions

We have developed a simple method for determining the best mapping arrangement to achieve a specific design goal without encountering the potential challenges of the binary linear optimization algorithm. This process helps us find the most efficient way to distribute an artificial neural network across the optimal number of physical tiles with a given capacity. By making certain assumptions about the scalability of the peripheral circuits associated with the tile, we can estimate the total tile area. To calculate the chip area, we must add additional components such as control units, IO components, and inter-tile communication. While our method provides an optimal tile array capacity and the necessary number of tiles, we must also consider other constraints. For instance, an increase in the number of tiles will lead to greater complexity in inter-tile communication. We have observed that by allowing for a change in the aspect ratio of the tile array, we can reduce the number of tiles while maintaining a roughly constant total tile area. This finding applies to inference. However, when it comes to training, it can be argued that a square tile is beneficial, as it allows for the use of a single set of ADCs in forward and backward paths, thereby minimizing the tile area. Future research could explore the impact of manufacturing yield on the optimization process, which would impose additional constraints on the optimal tile array capacity, or introduce constraints related to tile communication.

## Acknowledgment

We thank Tayfun Gokmen and Supratik Guha for valuable feedback on the manuscript. This work was supported in part by the U.S. Department of Energy, Office of Science, for support of microelectronics research, under contract number DE-AC0206CH11357, and in part by The Physical Science Department of Argonne National Lab.